\documentclass[10pt,journal,compsoc]{IEEEtran}
\ifCLASSOPTIONcompsoc
  \usepackage[nocompress]{cite}
\else
  \usepackage{cite}
\fi
\ifCLASSINFOpdf
  \usepackage[pdftex]{graphicx}
\else
\fi
\begin{document}
\title{Age and Gender Classification from Ear Images}
\author{\IEEEauthorblockN{Dogucan Yaman\footnotemark\textsuperscript{1, *}, Fevziye Irem Eyiokur\textsuperscript{1, *}, Nurdan Sezgin\textsuperscript{2}, Haz{\i}m Kemal Ekenel\textsuperscript{1}} 

\IEEEauthorblockA{\textsuperscript{1}Department of Computer Engineering, \textsuperscript{2}Department of Medical Services and Techniques \\
\textsuperscript{1}Istanbul Technical University, \textsuperscript{2}Istanbul Ayd{\i}n University \\
Email: \{yamand16, eyiokur16, ekenel\}@itu.edu.tr, nurdansezgin@aydin.edu.tr } }

\markboth{Accepted for IAPR/IEEE IWBF 2018}
{}
\IEEEcompsoctitleabstractindextext{%

\begin{abstract}
In this paper, we present a detailed analysis on extracting soft biometric traits, age and gender, from ear images. Although there have been a few previous work on gender classification using ear images, to the best of our knowledge, this study is the first work on age classification from ear images. In the study, we have utilized both geometric features and appearance-based features for ear representation. The utilized geometric features are based on eight anthropometric landmarks and consist of 14 distance measurements and two area calculations. The appearance-based methods employ deep convolutional neural networks for representation and classification. The well-known convolutional neural network models, namely, AlexNet, VGG-16, GoogLeNet, and SqueezeNet have been adopted for the study. They have been fine-tuned on a large-scale ear dataset that has been built from the profile and close-to-profile face images in the Multi-PIE face dataset. This way, we have performed a domain adaptation. The updated models have been fine-tuned once more time on the small-scale target ear dataset, which contains only around 270 ear images for training. According to the experimental results, appearance-based methods have been found to be superior to the methods based on geometric features. We have achieved 94\% accuracy for gender classification, whereas 52\% accuracy has been obtained for age classification. These results indicate that ear images provide useful cues for age and gender classification, however, further work is required for age estimation. 
\end{abstract}
}

\maketitle

\renewcommand{\thefootnote}{\Alph{footnote}}
\footnotetext{\noindent\rule{8cm}{0.4pt}}
\footnotetext{\textsuperscript{*}\textit{The authors have equally contributed.}}
\footnotetext{\textit{\textcopyright\textcopyright2018 IEEE. Personal use of this material is permitted. Permission from IEEE must be obtained for all other uses, in any current or future media, including reprinting/republishing this material for advertising or promotional purposes, creating new collective works, for resale or redistribution to servers or lists, or reuse of any copyrighted component of this work in other works.}}

\IEEEdisplaynotcompsoctitleabstractindextext

\IEEEpeerreviewmaketitle

\section{Introduction}\label{sec1}
Ear biometrics has become a popular research topic \cite{emervsivc2017ear}. A recent challenge, named as Unconstrained Ear Recognition Challenge \cite{emervsivc2017unconstrained} has shown the difficulties of performing person identification from ear images in the wild. To complement the identity related information from ear images, utilizing soft biometric traits, such as age and gender information can be auxiliary. For this purpose, in this paper, we have extensively investigated the tasks of age and gender classification from ear images. 

Biometric characteristics are expected to not change much over time, easy to obtain and unique for each individual \cite{kumar2012automated}. Because of its several features, ear is an important modality in biometric studies and forensic science for identification. For example, compared to facial appearance, which is influenced by changes in facial expression, facial hair or makeup, ear appearance is relatively constant. Auricular is also a defining feature of face \cite{yavuz27investigation}. Among the ear parts, earlobe is the most frequently used part in forensic cases. It is the only part of the ear that continues to grow and changes their shape \cite{nixon2010use}. Ear can be still visible in whole or partly covered face in the captured images from security cameras, and can be used as an auxiliary information for identification. Also, when faces are viewed in profile, ear can be easily captured from video recordings or photos \cite{abaza2013survey}.

Although, there have been many studies on using ear images for person identification \cite{emervsivc2017ear, abaza2013survey}, the number of studies on extracting soft biometric traits, such as age and gender, from ear images is limited. To the best of our knowledge, this study is the first work on age classification from ear images. However, there have been a couple of previous work on using ear images for gender classification \cite{gnanasivam2013gender,zhang2011hierarchical,khorsandi2013gender,lei2013gender}. In \cite{gnanasivam2013gender}, the ear-hole is used as the reference point for the measurements. The Euclidean distances between ear hole and seven features of ear, which are identified from masked ear images, are calculated. They used an internal database, which has 342 samples, for the experiments. They have employed Bayes classifier, KNN classifier, and neural networks. The best performance is achieved by KNN with 90.42\% classification accuracy. In~\cite{zhang2011hierarchical}, profile face images and ear images are used separately and are classified by support vector machines (SVM) with histogram intersection kernel. They performed score level fusion based on Bayesian analysis to improve the accuracy. The 2D images of UND biometrics dataset collection~F \cite{yan2005empirical} have been used for the experiments. Fusion leads to 97.65\% accuracy, whereas face only performance is around 95.43\% and ear only accuracy is around 91.78\%. In \cite{khorsandi2013gender}, Gabor filters have been utilized to extract features and classification has been performed with extracted features based on dictionary learning. The dictionary has been built from training samples and used in the test phase to represent a test sample as a linear combination of the training data. UND biometric dataset collection~J \cite{yan2005empirical}, which contains large appearance, pose, and illumination variability, has been used in the experiments. The best obtained accuracy reported in the paper is 89.49\% has been achieved by using 128 features. In \cite{lei2013gender}, gender classification is performed both on 2D and 3D ear images. 3D ears are automatically detected and aligned. The experiments were performed on UND dataset collections F and J2 \cite{yan2005empirical}. Histogram of Indexed Shapes features were extracted and classified by SVM. The average performance of the system was 92.94\%.

%Also law enforcement can identify criminals using identification methods such as DNA profiling, but the cost of collection and the reliability of the data are changing from country to country. Ear-related issues cheaper than DNA and more reliable evidence in court as a significant biometrics because of its relatively undestroyable nature, compared to DNA or not being able to present accidentally at the crime scene [Referans bulunamadı!].

%Ear biometrics is an attractive subject because of its rich and stable structure from birth, almost unique, and does not change with facial expression and pose changing \cite{kumar2012automated}. Face can change with cosmetic, beard/mustache, and hairstyle. However, face also changes with emotions such as sadness, happiness, fear, or surprise. İn contrast, the ear is relatively constant and unchanging characteristics of a structure. Ear is also smaller than face. This indicates that low-resolution images could be studied faster and more effective \cite{choras2010ear}. 
%Facial image analysis is an important area of forensic identification. 

%Human identification is very important in the forensic sciences \cite{roelofse2008photo}. 

In this paper, we present an extensive analysis on age and gender classification from ear images. We have explored use of both geometric features and appearance-based features for ear representation. Geometric features are based on eight landmarks determined on the ear. From these landmarks, to extract the features, we have calculated 14 different distances between them as well as performed two area calculations. To classify these extracted features, four different classifiers ---logistic regression, random forests, support vector machines, neural networks--- have been employed. The appearance-based methods are based on well-known deep convolutional neural network (CNN) models, namely, AlexNet \cite{krizhevsky2012imagenet}, VGG-16 \cite{VGG2014}, GoogLeNet \cite{GoogLeNet2015}, and SqueezeNet \cite{iandola2016squeezenet}. They have been fine-tuned twice, first on a large-scale ear dataset to provide domain adaptation, then on the small-scale target ear dataset. In the experiments, appearance-based methods have outperformed geometric feature-based methods. We have achieved 94\% accuracy for gender classification, exceeding the attained accuracies in the previous studies. For age classification 52\% accuracy has been obtained. In summary, the contributions of the paper can be listed as below:

\begin{itemize}
\item We have explored geometric and appearance-based features for age and gender classification from ear images.
\item For geometric features, we have used eight landmark points on the ear and derived 16 features from them.
\item For appearance-based methods, we have utilized a large-scale ear dataset \cite{yaman2017domain}, which has been built from the profile and close-to-profile face images in the Multi-PIE face dataset \cite{gross2010multi}. This way, we have efficiently transferred and benefited from the well-known CNN models for the problem at hand.
\item We have achieved superior performance for gender classification compared to the previous work. We have presented the first work on age classification from ear images.
\end{itemize}

The remainder of the paper is organized as follows. In Section II, we explain the geometric features, the classifiers used with them, and the convolutional neural networks used for ear appearance representation and classification. In Section III, we introduce the dataset and experimental setup, and present the obtained results. Finally, in Section IV, we conclude the paper and point the future research directions.

\section{Methodology}

In this section, we present the utilized geometric features and the employed classifiers on them, as well as the appearance-based representation and classification.

\subsection{Geometric Features}

Identified landmarks and performed measurements are shown in Fig. \ref{fig_sim}. The geometric features calculated from these landmarks are listed in Table \ref{table_faetures}. In Table \ref{table_faetures}, \textit{Selected} column refers to the features that were found important by the random forest classifier. In summary, we have used 8 landmark points and calculated 16 measurements from them to generate the feature vector. To calculate the rectangular area of the ear, we used the most outer point between Obs and Obi on the left, Sa on the upper side, Pa on the right, and Sba at the bottom. To calculate the polygon area of the ear, Obs, Sa, Pa, Sba, Obi, and T points were used. The remaining measurements are distances between two landmarks as listed in Table \ref{table_faetures} \cite{sezgin2016}. Definition of the utilized landmarks are as below:

\begin{figure}[t]
\centering
\includegraphics[width=3.5in]{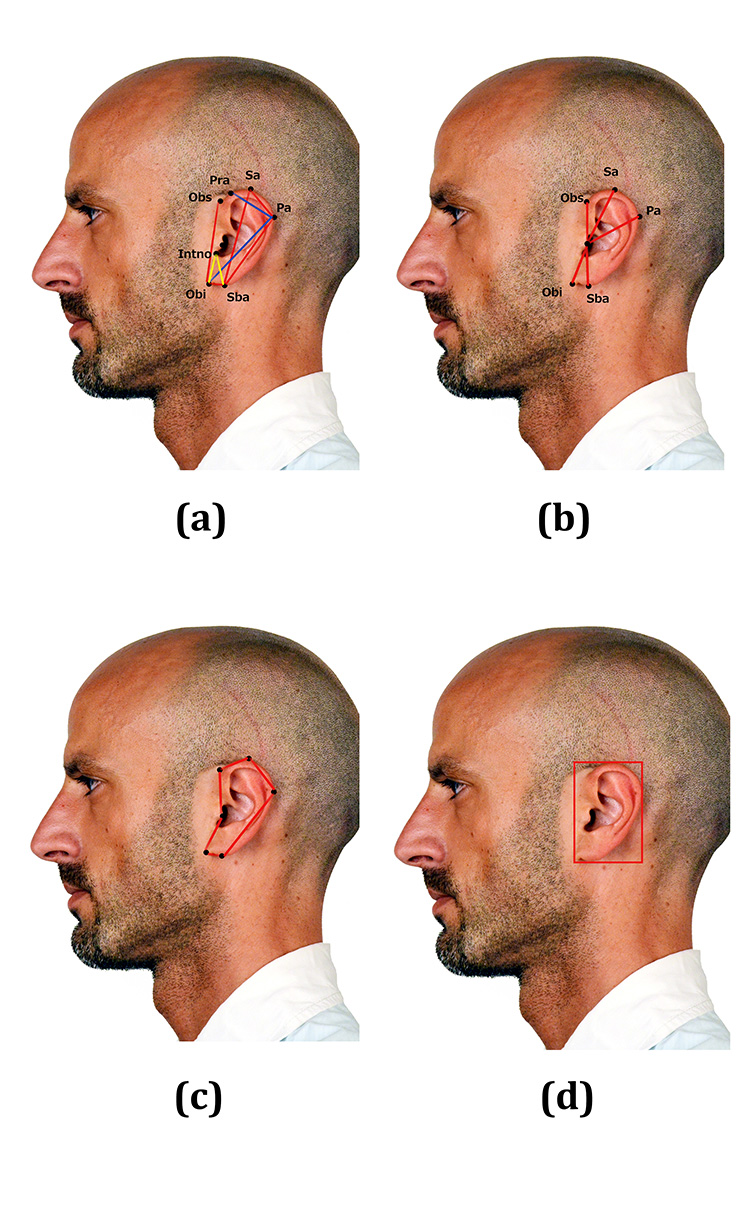}
\caption{Used landmarks and measurements for geometric features. (a)-(b) Landmarks and calculated distances, (c)-(d) Calculated areas.}
\label{fig_sim}
\end{figure}

\begin{itemize}
\item \textit{Otobasion superius (Obs)} \cite{Guyomarc2012}: It is the point, which connects helix in the temporal region, and it determines the upper limit of the junction of the ear with the face.
\item \textit{Otobasion inferius (Obi)} \cite{Guyomarc2012}: This is the connecting point of the earlobe to the cheek. It determines the lower bound of the junction of the ear with the face. 
\item \textit{Tragus (T)} \cite{Menezes2009}: Tragus is a protruding part in front of the hearing canal.
\item \textit{Superaurale (Sa)} \cite{Menezes2009}: It is the highest point of the auricular.
\item \textit{Subaurale (Sba)} \cite{Menezes2009}: It is the lowest point of the auricular. This point varies person to person. It is the commissure point at the bottom of the ear with face skin on the person who does not have an earlobe (attached earlobes).
\item \textit{Postaurale (Pa)} \cite{Menezes2009}: It is the most outer point on the ear curve to the back.
\item \textit{Preaurale (Pra)} \cite{Menezes2009}: It is the front side of the ear, and present at the helix added level of the head.
\item \textit{Intertragic notch (Intno)} \cite{hurley2008}: It is the deep notch between the tragus and antitragus. 
%\item \textit{Rectangular area of the ear}:
%In the front side of the ear, used the most outer point between Obs and Obi, on the upper Sa, on the back Pa, at the bottom Sba points used as a border, and the ear placed into the frame and the are calculated.

%\subsubsection{Polygons area of the ear}
%Obs, Sa, Pa, Sba, Obi and T points were used for this calculating. The obtained polygon that drawn line between points were calculated (tez).

\end{itemize}

\begin{table}[t]
\renewcommand{\arraystretch}{1.3}
\caption{List of Geometric Features}
\label{table_faetures}
\centering
\begin{tabular}{|c||c|}
\hline
%Method & \multicolumn{2}{c}{Age Classification} & \multicolumn{2}{c}{Gender C.} \\
Features & Selected \\
\hline
Obs-Obi & - \\
\hline
Sa-Sba & - \\
\hline
Sa-Pa & - \\
\hline
Pa-Sba & - \\
\hline
Obi-Sba & + \\
\hline
Obi-Pa & - \\
\hline
T-Obs & - \\
\hline
T-Sa &  + \\
\hline
T-Pa & - \\
\hline
T-Sba & +  \\
\hline
T-Obi & - \\
\hline
Pra-Pa & - \\
\hline
Intno-Obi & +  \\
\hline
Intno-Sba & +  \\
\hline
Ear Rectangle Area & +  \\
\hline
Ear Polygon Area & - \\
\hline
\end{tabular}
\end{table}

Since each geometric feature has a different value range, in order to normalize them, for each feature we have calculated its mean value and the standard deviation on the training set. Then, we have normalized them so that they have zero mean and unit variance. 

Using random forest classifier, the importance of each feature was measured with respect to the predictability of the correct output variable. According to these importances, we determined a threshold value to choose the features. This way, six of the 16 geometric features of the ear were selected. We have observed that, since the amount of available data were limited, this feature selection scheme has improved the results.  %Further, experiments in this study were carried out with these features which were transformed into a vector. When these experiments have been performed, the geometric features were normalized with the feature-based as mean value being zero.

\subsection{Classification of Geometric Features}

In this section, the classifiers which were used with geometric features, are explained and information about their parameters is presented. These classifiers are logistic regression, random forest, support vector machines, and neural networks. % are used for training and classifying geometric features for age and gender classification from ear.

\subsubsection{Logistic Regression}
Logistic regression provides a linear discrimination between different classes. As the name implies, its objective is to  minimize the logistic function. Due to limited amount of training data, in the experiments, we used logistic regression with L2 regularization. 

%, input values are combined with weights as linearly to predict an output value. In multi class classification, logistic regression predicts some results with probabilities like linear regression and these predictions are transformed to the class values with using logistic function. This is the most important difference between linear regression and logistic regression. In experiment with logistic regression, we have used l2 regularization. %and we obtained the best result with regularization strength. 

\subsubsection{Random Forest}
Random forest is an ensemble machine learning algorithm and was used for classification as well as feature selection in this study. Random forest works with sub-trees that are learned on a part of the training data. The advantage of using sub-trees is that each sub-tree's predictions has lower correlation. Besides, each sub-tree uses different samples from training set instead of using all of them. In the end, the average result is calculated from different sub-trees. In our experiments, we obtained the best result using 1000 sub-trees in the forest. %Further, we utilized from random forest to select some features which are more important.

\subsubsection{Support Vector Machines}
Support vector machine \cite{SVM1995} is a classifier, which finds a decision boundary between two classes that enforces a margin. In some cases, different classes cannot be separated linearly, then nonlinear kernels are employed. In gender classification problem, we used binary SVM classifier but in age classification, there exists more than one class. For that case, we applied one-vs-one scheme. In the experiments, \textit{radial basis function (rbf)} kernel were used. Gamma value was set to 1/number of features and the penalty parameter was set to \textit{C=250}. These values have been determined empirically according to the accuracies obtained on the validation set. %, and \textit{radial basis function (rbf)} kernel with kernel coefficient gamma equals to 1/number of features.

\subsubsection{Neural Network}
We have employed a neural network that contains 3 hidden layers. This parameter was determined again emprically according to the accuracies obtained on the validation set. As we increased the number of layers, we observed that the training accuracy increased and the validation accuracy decreased, which is an indication of \textit{overfitting} due to limited amount of data. 

%Input layer of neural network has neurons for each input and output layer contains neurons with number of output dimensions which is 5 for age classification and 2 for gender classification. Further, there are some non-linear layers which is named as hidden layer, between input layer and output layer. Each neuron of the hidden layer transforms the values which is from the previous layer, using linear equations with non-linear activation function. In this study, we obtained the best result with neural network that contains 3 hidden layers. As we increased the number of layers, we observed that the training accuracy increased and the test accuracy decreased. In other words, as the amount of data is small, when we add more layers and neurons, artificial neural network would be more complicated and overfitting would be observed. Further, we used \textit{ReLU} \cite{krizhevsky2012imagenet} activation function for hidden layers and \textit{lbfgs} optimizer.

\subsection{Appearance-based Representation and Classification}
In this study to represent and classify the ear appearance, we have employed convolutional neural networks. We have benefited from well-known CNN architectures, namely, AlexNet~\cite{krizhevsky2012imagenet}, VGG-16 \cite{VGG2014}, GoogLeNet \cite{GoogLeNet2015} and SqueezeNet~\cite{iandola2016squeezenet}. At the end of all these deep network architectures, \textit{softmax} layer is used as classifier.

The first deep convolutional neural network architecture used in this study is AlexNet \cite{krizhevsky2012imagenet}. It is one of the most popular CNN architectures as the winner model of ILSVRC 2012 challenge. AlexNet \cite{krizhevsky2012imagenet} contains only five convolutional layers and three fully connected layers. Therefore, it is relatively a less deep architecture. In the network training, to prevent overfitting dropout method \cite{srivastava2014dropout} has been used.

VGG architecture \cite{VGG2014} has two versions, VGG-16 which was used in this study, and VGG-19. VGG-16 \cite{VGG2014} contains 16 convolutional layers, 3 fully connected layers and softmax classifier after convolutional layers as in AlexNet~\cite{krizhevsky2012imagenet}. The main difference between AlexNet~\cite{krizhevsky2012imagenet} and VGG-16~\cite{VGG2014} is that VGG-16~\cite{VGG2014} is a deeper network than AlexNet~\cite{krizhevsky2012imagenet} and it uses many small size filters. 

GoogLeNet \cite{GoogLeNet2015} is a deeper network and contains 22 layers. It is based on the inception module and is mainly a concatenation of several inception modules. The inception module contains several filters of different sizes. Different filtering outputs are combined and this way multiple features from the input data are extracted. The architecture is also efficient in terms of the number of parameters. Although, it is deeper than the AlexNet \cite{krizhevsky2012imagenet}, it has about twelve times fewer parameters.

The last CNN architecture, which is SqueezeNet~\cite{iandola2016squeezenet}, proposed a new approach to reduce the number of parameters and model size. $1 \times 1$ filters are used rather than $3 \times 3$ filters. This architecture also contains residual connections to increase efficiency of back-propagation learning. In addition, there is no fully connected layers. Average pooling layer is used instead of fully connected layers. %In the end, softmax layer is utilized as classifier.

\begin{figure}[t]
\centering
\includegraphics[width=3.4in]{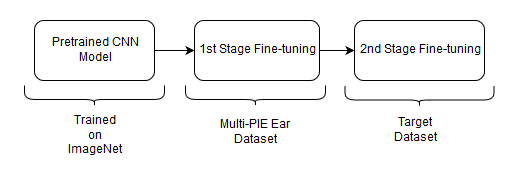}
\caption{Visualization of two-stage finetuning approach. }
\label{fig_network}
\end{figure}

\begin{table}[t]
\renewcommand{\arraystretch}{1.3}
\caption{Age and Gender Distribution}
\label{table_distribution}
\centering
\begin{tabular}{|c||c||c|}
\hline
%Method & \multicolumn{2}{c}{Age Classification} & \multicolumn{2}{c}{Gender C.} \\
Age Group & Female & Male \\
\hline
18-28 & 35 & 36 \\
\hline
29-38 & 32 & 43 \\
\hline
39-48 & 27 & 45 \\
\hline
49-58 & 27 & 33 \\
\hline
59-68+ & 29 & 31 \\
\hline
\end{tabular}
\end{table}

\begin{figure}[t]
\centering
\includegraphics[width=3.2in]{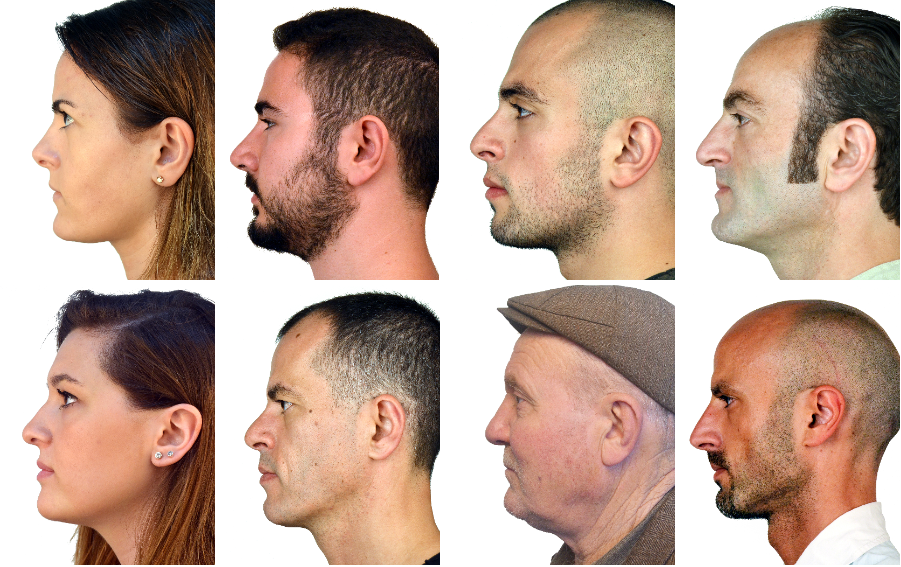}
\caption{Profile images belong to different subjects from our dataset.}
\label{fig_dataset}
\end{figure}

In this study, we have first taken pretrained AlexNet \cite{krizhevsky2012imagenet}, VGG-16 \cite{VGG2014}, GoogLeNet \cite{GoogLeNet2015} and SqueezeNet \cite{iandola2016squeezenet} models, which were trained on the ImageNet dataset \cite{imagenet2014}. Then, we have applied fine-tuning on the ear datasets. In our previous works on age and gender classification from face images \cite{ozbulak2016transferable} and on person identification from ear images \cite{yaman2017domain}, we have shown that transferring a pretrained deep CNN model from a closer domain leads to improved performance. That is, for age and gender classification from face images transferring a pretrained model that were trained on face images and for person identification from ear images transferring a pretrained model that were trained on ear images is better than transferring a pretrained model that were trained on generic object images, such as the ones from ImageNet dataset \cite{imagenet2014}. With this finding in mind, we have applied two-stage fine-tuning as shown in Fig. \ref{fig_network}. In the first stage, we have fine-tuned the pretrained CNN models on a large ear dataset to provide domain adaptation. In the second stage, we have performed once more fine-tuning, but this time using our ear dataset that contains age and gender labels. We obtained these ear images from profile face images in Fig. \ref{fig_dataset}. For the first stage, we have used the Multi-PIE ear dataset~\cite{yaman2017domain} that were prepared by processing the Multi-PIE face dataset \cite{gross2010multi}. It is currently the largest ear dataset containing 17183 ear images from 205 different subjects. This dataset has been created by running an ear detector on the profile and close-to-profile face images available in the Multi-PIE dataset~\cite{gross2010multi}.

In all training steps, the learning rate of the last fully connected layer of AlexNet \cite{krizhevsky2012imagenet}, VGG-16 \cite{VGG2014}, and GoogLeNet~\cite{GoogLeNet2015} has been increased by ten times. Increasing the learning rates of last layers during fine-tuning is a typical approach to improve the accuracy of the classification, since these layers focus more on high-level features and classification. The output of \textit{softmax} layer has adjusted to the number of classes in all models. Global learning rate has been selected as 0.0001 for all models except SqueezeNet \cite{iandola2016squeezenet}, for which we set it to 0.0004. %in SqueezeNet \cite{iandola2016squeezenet}.}

\section{Experimental Results}

Our dataset contains profile face images of 338 different subjects. All subjects in this dataset are over 18 years old. Sample images from the dataset can be seen from Fig. \ref{fig_dataset}. These subjects are categorized into five different age groups. These age groups are 18-28, 29-38, 39-48, 49-58, 59-68+, respectively. Age groups are categorized with respect to the changes in geometric features. Geometric measurements are relatively closer to each other at the specified age ranges. As listed in Table \ref{table_distribution}, distribution of subjects over different age groups is relatively even. Among the 338 subjects, 188 subjects are males and 150 of them are females. The dataset contains just one profile image for each subject. OpenCV~\cite{opencv} ear detection implementation has been utilized to detect and crop ear regions from profile face images and false detections have been eliminated manually. The dataset has been divided into two parts as training set and test set. 80\% of the dataset has been utilized for training and 20\% has been employed for testing. Besides, the validation set has been selected from training data for validation purposes during training. Distribution of training and testing sets is given in Table~\ref{table_setup}. To prevent unbalanced class distribution, each age and gender group of the dataset has been divided with respect to train-test division procedure (80-20\%), separately. Besides, the training and test sets do not contain the same subjects, that is, the experiments have been conducted in subject-independent manner.

%Because there is only one image for each subject on our dataset.  %The validation set has been built from training data and 5-fold cross validation has been performed. Age distribution for age classification experiments and gender distribution for gender classification experiments are considered, while dataset is divided into train and test. Finally, the contributions of domain adaptation and two-stage fine-tuning are presented in below sections.

\begin{table}[t]
\renewcommand{\arraystretch}{1.3}
\caption{Experimental Setup}
\label{table_setup}
\centering
\begin{tabular}{|c||c||c||c|}
\hline
%Method & \multicolumn{2}{c}{Age Classification} & \multicolumn{2}{c}{Gender C.} \\
Group & Train Data & Validation Data & Test Data \\
\hline
18-28 & 50 & 7 & 14\\
\hline
29-38 & 53 & 7 & 15\\
\hline
39-48 & 51 & 7 & 14\\
\hline
49-58 & 42 & 6 & 12\\
\hline
59-68+ & 42 & 6 & 12\\
\hline
Male & 131 & 19 & 38\\
\hline
Female & 105 & 15 & 30\\
\hline
\end{tabular}
\end{table}
%\begin{table}[htbp]
%\renewcommand{\arraystretch}{1.3}
%\caption{Experimental Setup}
%\label{table_setup}
%\centering
%\begin{tabular}{|c||c||c|}
%\hline
%%Method & \multicolumn{2}{c}{Age Classification} & \multicolumn{2}{c}{Gender C.} \\
%Set & Age Classification & Gender Classification \\
%\hline
%Train data & 269 & 272 \\
%\hline
%Test data & 69 & 66 \\
%\hline
%\end{tabular}
%\end{table}

In the experiments, ear images have been resized to $256~\times~256$ pixel resolution for the deep CNN models. During training, five different crops, with $224 \times 224$ resolution for VGG-16~\cite{VGG2014} and GoogLeNet~\cite{GoogLeNet2015} and $227~\times~227$ resolution for AlexNet~\cite{krizhevsky2012imagenet} and SqueezeNet~\cite{iandola2016squeezenet} have been taken from these $256~\times~256$ pixel resolution images. During testing, single crop is taken from the center of the image at the appropriate size with respect to the used architecture.

The number of images in our dataset is limited to train deep convolutional neural network models. To overcome this limitation and increase the amount of data, we have applied data augmentation and obtained 55 different images with different variations from each image. We have also performed data augmentation on the Multi-PIE ear dataset. In this study for data augmentation, we have utilized Imgaug tool\footnote{http://github.com/aleju/imgaug}. Many images have been created by different transformation techniques. First, flipped version of the original image has been created. Then, images with different brightness levels have been generated by adding some positive and negative values to the pixels' intensity values. These positive and negative values are in the range of [-55 +55] with an increment by ten. The second way of changing brightness levels of the images have been applied by multiplying pixels' intensity values with some constant values. These constant values have been chosen between 0.5 and 1.5 by incrementing by 0.1. To improve the generalization of the deep CNN models, we have performed Gaussian blur and dropout. For Gaussian blur, we have produced blurred images at different levels by using different sigma values, such as 0.25, 0.5, 0.75, 1, 1.25, 1.5, and 2. For dropout, some pixels have been dropped and new noisy images have been created. The last augmentation method, sharpening, has been applied on each image by choosing values between 0.5 and 2.0 by increasing by steps of 0.1. After all these processes, we have obtained 14795 training images for age classification and 14960 training images for gender classification.

\begin{table}[!t]
\renewcommand{\arraystretch}{1.3}
\caption{Gender classification results}
\label{table_gender}
\centering
\begin{tabular}{|c||c|}
\hline
Approach & Test Result \\
\hline
Logistic Regression & 47\% (Geometric) \\
\hline
Random Forest & 47\% (Geometric) \\
\hline
SVM & 65\% (Geometric) \\
\hline
3 hidden layers NN & 58\% (Geometric) \\
\hline
AlexNet & 88\% (Appearance) \\
\hline
VGG-16 & 89\% (Appearance) \\
\hline
GoogLeNet & 94\% (Appearance) \\
\hline
SqueezeNet & 89\% (Appearance) \\
\hline
\end{tabular}
\end{table}

\subsection{Gender Classification Results}
Gender classification results are presented in Table \ref{table_gender}. In the table, first column includes the name of the classifier and the second one contains the corresponding classification accuracy. To remind the reader about the used features, the type of the features are included in parenthesis in the second column. As can be seen from the table, appearance-based approaches are superior to the classifiers that utilize geometric features. Considering that the chance level of correct gender classification is 50\%, the results obtained by using the geometric features are very poor. One main reason for this inferior performance could be the normalization step that has been applied on the geometric features. During the normalization procedure ---making the features have zero mean and unit variance--- discriminative information about gender might have been lost. Therefore, the effect of normalization requires further analysis. %A possible solution could be performing a gender-based normalization, that is, to calculate separate mean and variance values for males and females. 
The appearance-based approaches have achieved around 90\% accuracy. The best performance has been obtained using the GoogLeNet architecture \cite{GoogLeNet2015} with 94\% correct classification. This accuracy exceeds the gender classification accuracies achieved in previous studies on gender classification from ear images~\cite{gnanasivam2013gender,zhang2011hierarchical,khorsandi2013gender,lei2013gender}. A comparison of these approaches are given in Table \ref{table_comparison}. Overall, in compliance with the findings of the previous work, we have found that ear images provide useful information to classify genders of the subjects.

\begin{table}[t]
\renewcommand{\arraystretch}{1.3}
\caption{Comparison of Gender Classification Methods}
\label{table_comparison}
\centering
\begin{tabular}{|c||c||c||c|}
\hline
Method & Dataset & No. of Img. & Accuracy \\
\hline
SVC \cite{zhang2011hierarchical} & UND Collection F & 942 & 91.7\% \\
\hline
Majority Voting \cite{khorsandi2013gender} & UND Collection J2 & 2430 & 89.49\%  \\
\hline
SVM \cite{lei2013gender} & UND Collection J2 & 2430 & 91.92\% \\
\hline
SVM \cite{lei2013gender} &  UND Collection F & 942 & 92.94\%   \\
\hline
KNN \cite{gnanasivam2013gender} & Internal dataset & 342 & 90.42\% \\
\hline
CNN & Our Dataset & 338 & 94\%  \\
\hline
\end{tabular}
\end{table}

\subsection{Age Classification Results}
%Age classification experiments have been obtained from geometric features and pixel-wise ear images. Same classifiers and deep CNN models in gender classification experiments have been employed for age classification. Further, in deep CNN approach, domain adaptation with two-stage fine-tuning and data augmentation have been applied and they improved the classification performance of the models.

Age classification results are presented in Table \ref{table_age}. The first column includes the name of the classifier and the second one contains the corresponding classification accuracy. To remind the reader about the used features, the type of the features are included in parenthesis in the second column. This time performance gap between the geometric feature-based methods and appearance-based methods is close. However, appearance-based methods have been found again superior. Using geometric features, the best performance is achieved with 3 hidden layer neural network and logistic regression, reaching 43\% accuracy. The best performance has been obtained with the appearance-based method using the GoogLeNet architecture~\cite{GoogLeNet2015} with 52\% correct classification. Compared to the performance achieved for gender classification, age classification accuracy is relatively low. One possible reason for this outcome is the limited amount of samples per each age group. Since the number of classes is higher in age classification, the amount of samples per class is less. We plan to extend the dataset and analyze the results further. Since the accuracies obtained by the geometric feature-based methods and appearance-based methods are close, combining these two approaches could be another way to improve the performance. Overall, appearance provides more information compared to geometric features, therefore, have been found to be more useful for age and gender classification. 

\begin{table}[!t]
\renewcommand{\arraystretch}{1.3}
\caption{Age estimation results}
\label{table_age}
\centering
\begin{tabular}{|c||c|}
\hline
Approach & Test Result \\
\hline
Logistic Regression & 43\% (Geometric) \\
\hline
Random Forest & 34\% (Geometric)\\
\hline
SVM & 39\% (Geometric) \\
\hline
3 hidden layers NN & 43\% (Geometric)\\
\hline
AlexNet & 45\% (Appearance) \\
\hline
VGG-16 & 39\% (Appearance) \\
\hline
GoogLeNet & 52\% (Appearance) \\
\hline
SqueezeNet & 39\% (Appearance) \\
\hline
\end{tabular}
\end{table}

\section{Conclusion}
In this paper, we have presented a thorough study on age and gender classification from ear images. To the best of our knowledge, this study is the first work on age classification from ear images and one of the few studies on gender classification using ear images. In the study, we have employed both geometric features and appearance-based features for ear representation. The geometric features are calculated with respect to eight anthropometric landmarks on the ear and consist of 14 distance measurements and two area calculations. These features have been then classified using four different methods: logistic regression, random forests, support vector machines, and neural networks. The appearance-based methods are based on deep convolutional neural networks. The well-known CNN models, namely, AlexNet \cite{krizhevsky2012imagenet}, VGG-16~\cite{VGG2014}, GoogLeNet \cite{GoogLeNet2015}, and SqueezeNet \cite{iandola2016squeezenet} have been adopted for the study. To transfer them efficiently to the task at hand, they have been first fine-tuned on a large-scale ear dataset that has been built from the profile and close-to-profile face images available in the Multi-PIE face dataset~\cite{gross2010multi}. Afterwards, the updated models have been fine-tuned again on the small-scale target ear dataset. As a result of the experiments, appearance-based methods have been found to be superior to the methods based on geometric features. We have achieved 94\% accuracy for gender classification, whereas 52\% accuracy has been obtained for age classification. These results indicate that ear images provide useful cues for age and gender classification. However, gender classification using geometric features require further work. It has been noticed that for gender classification geometric features are sensitive to the normalization. Therefore, better normalization schemes have to be explored. For age estimation, we believe that the main reason for the lower performance is the lack of sufficient amount of training samples from each age group. We plan to extend the dataset and train the age classification system with larger amount of samples. We also aim to make comparisons by performing experiments on popular datasets, such as UND-F and UND-J2 \cite{yan2005empirical}. Besides, we also plan to investigate the complementarities between the geometric and appearance-based features. Moreover, we plan to combine profile face images and ear images for age and gender classification.


\section*{Acknowledgment}

This work was supported by Istanbul Technical University Research Fund, ITU BAP, Project No. MGA-2017-40893.

\bibliographystyle{IEEEtran}
\bibliography{refs}

\end{document}